\documentclass[final]{cvpr}

\usepackage{times}
\usepackage{epsfig}
\usepackage{graphicx}
\usepackage{amsmath}
\usepackage{amssymb}

\usepackage{paralist}
\usepackage{enumitem}

\usepackage{multirow}
\usepackage{makecell}
\usepackage{booktabs}
\usepackage{tabularx}

\usepackage{url}

\usepackage[ruled,vlined]{algorithm2e}

\usepackage{subcaption}

\usepackage[pagebackref=true,breaklinks=true,colorlinks,bookmarks=false]{hyperref}

\def\cvprPaperID{6737} 
\def\confYear{CVPR 2021}

\begin{document}

\title{UAV-Human: A Large Benchmark for Human Behavior Understanding with Unmanned Aerial Vehicles}



\author{Tianjiao Li{$^{1}$} \qquad Jun Liu{$^{1}$} 
\qquad Wei Zhang{$^{2}$} 
\qquad Yun Ni{$^{1}$} \qquad Wenqian Wang{$^{2}$} \qquad Zhiheng Li{$^{2}$} \\
\normalsize {$^{1}$} Information Systems Technology and Design, Singapore University of Technology and Design, Singapore\\
\normalsize {$^{2}$} School of Control Science and Engineering, Shandong University, Jinan, Shandong\\
{\tt\small tianjiao\_li@mymail.sutd.edu.sg, \{jun\_liu, ni\_yun\}@sutd.edu.sg}\\
{\tt\small davidzhang@sdu.edu.cn, \{wqwang, zhihengli\}@mail.sdu.edu.cn}
}

\maketitle
\pagestyle{empty}
\thispagestyle{empty}

\begin{abstract}
Human behavior understanding with unmanned aerial vehicles (UAVs) is of great significance for a wide range of applications, which simultaneously brings an urgent demand of large, challenging, and comprehensive benchmarks for the development and evaluation of UAV-based models. 
However, existing benchmarks have limitations in terms of the amount of captured data, types of data modalities, categories of provided tasks, and diversities of subjects and environments. 
Here we propose a new benchmark \-- UAV-Human \-- for human behavior understanding with UAVs,
which contains 67,428 multi-modal video sequences and 119 subjects for action recognition, 22,476 frames for pose estimation, 41,290 frames and 1,144 identities for person re-identification, and 22,263 frames for attribute recognition.
Our dataset was collected by a flying UAV in multiple urban and rural districts in both daytime and nighttime over three months,
hence covering extensive diversities w.r.t subjects, backgrounds, 
illuminations, 
weathers,
occlusions,
camera motions,
and UAV flying attitudes.
Such a comprehensive and challenging benchmark shall be able to promote the research of UAV-based human behavior understanding, including action recognition, pose estimation, re-identification, and attribute recognition.
Furthermore, 
we propose a fisheye-based action recognition method that mitigates the distortions in fisheye videos via learning unbounded transformations guided by flat RGB videos. Experiments show the efficacy of our method on the UAV-Human dataset. The project page: \href{https://github.com/SUTDCV/UAV-Human}{https://github.com/SUTDCV/UAV-Human}. 



\end{abstract}

\section{Introduction}
Given the flexibility and capability of long-range tracking, unmanned aerial vehicles (UAVs) equipped with cameras are often used to collect information from remote for the scenarios where it is either impossible or not sensible to use ground cameras \cite{nex2014uav,torresan2017forestry,mueller2016benchmark}. One particular area where UAVs are often deployed is human behavior understanding and surveillance in the wild, where video sequences of human subjects can be collected for analysis (such as action recognition, pose estimation, human re-identification, and attribute analysis), and for subsequent decision making. 

Compared to videos collected by common ground cameras,
the video sequences captured by UAVs generally present more diversified yet unique viewpoints, more obvious motion blurs, and more varying resolutions of the subjects, owing to the fast motion and continuously changing attitudes and heights of the UAVs during flight. These factors lead to significant challenges in UAV-based human behavior understanding, clearly requiring the design and development of human behavior understanding methods specifically taking the unique characteristics of UAV application scenarios into consideration. 






Existing works \cite{deng2009imagenet} have demonstrated the great importance of leveraging
large, comprehensive, and challenging benchmarks to develop and evaluate the state-of-the-art deep learning methods for handling various computer vision tasks. 
However, in the UAV-based human behavior understanding area, 
existing datasets \cite{barekatain2017okutama,perera2018uav} have limitations in multiple aspects, including: 
(1) Very limited number of samples, while a large scale of the dataset is often important for mitigating over-fitting issues and enhancing the generalization capability  of the models developed on it. 
(2) Limited number and limited diversity of subjects, while the diversities of human ages, genders, and clothing are crucial for developing robust models for analyzing the behaviors of various subjects. 
(3) Constrained capturing conditions. 
In practical application scenarios, UAVs often need to work in various regions (e.g., urban, rural, and even mountain and river areas),
under diversified weathers (e.g., windy and rainy weathers), in different time periods (e.g, daytime and nighttime). However, samples in existing datasets are usually collected under similar conditions, obviously simplifying the challenges in real-world UAV application scenarios. 
(4) Limited UAV viewpoints and flying attitudes. UAVs can experience frequent (and sometimes abrupt) position shifts during flying, which not only cause obvious viewpoint variations and motion blurs, but also lead to significant resolution changes. However, the UAVs in most of the existing datasets only present slow and slight motions with limited flying attitude variations.
(5) Limited types of data modalities.
In practical scenarios, we often need to deploy different types of sensors to collect data under different conditions. For example, infrared (IR) sensors can be deployed on UAVs for human search and rescue in the nighttime, while fisheye cameras are often used to capture a broad area. This indicates the significance of collecting different data modalities, to facilitate the development of various models for analyzing human behaviors under different conditions.
However, most of the existing UAV datasets provide conventional RGB video samples only.
(6) Limited categories of provided tasks and annotations.
As for UAV-based human behavior understanding, various tasks, such as action recognition, pose estimation, re-identification (ID), and attribute analysis, are all of great significance, which indicates the importance of providing thorough annotations of various tasks for a comprehensive behavior analysis. However,
most of the existing datasets provide annotations for one or two tasks only. 





The aforementioned limitations in existing datasets clearly show the demand of a larger, more challenging, and more comprehensive dataset for human behavior analysis with UAVs.
Motivated by this, in this work, we create UAV-Human, the first large-scale multi-modal benchmark in this domain. 
To construct this benchmark,
we collect samples by flying UAVs equipped with multiple sensors in both daytime and nighttime, over three different months, and across multiple rural districts and cities, which thus brings a large number of video samples covering extensive diversities w.r.t  human subjects, data modalities, capturing environments, and UAV flying attitudes and speeds, etc.

Specifically, a total of 22,476$\times$3 video sequences (consisting of three sensors: Azure DK, fisheye camera and night-vision camera) with 119 distinct subjects and 155 different activity categories are collected for action recognition; 22,476 frames with 17 major keypoints are annotated for pose estimation; 41,290 frames with 1,144 identities are collected for person re-identification; and 22,263 frames with 7 distinct characteristics are labelled for attribute recognition,
where the captured subjects present a wide range of ages (from 7 to 70) and clothing styles (from summer dressing to fall dressing). Meanwhile the capturing environments contain diverse scenes (45 sites, including forests, riversides, mountains, farmlands, streets, gyms, and buildings), different weather conditions (sunny, cloudy, rainy, and windy), and various illumination conditions (dark and bright).
Besides, different types of UAV flying attitudes, speeds, and trajectories are adopted to collect data, and thus our dataset covers very diversified yet practical viewpoints and camera motions in UAV application scenarios. 
Furthermore, the equipped different sensors enable our dataset to provide rich data modalities including RGB, depth, IR, fisheye, night-vision, and skeleton sequences.

Besides introducing the UAV-Human dataset, in this paper, we also propose a method for action recognition in fisheye UAV videos. Thanks to the wide angle of view,
fisheye cameras can capture a large area in one shot and thus
are often deployed on UAVs for surveillance. However, the provided wide angle in turn brings large distortions into the collected videos, making fisheye-based action recognition quite challenging. To mitigate this problem, we design a Guided Transformer I3D model to learn an unbounded transformation for fisheye videos under the guidance of flat RGB images. Experimental results show that such a design is able to boost the performance of action recognition using fisheye cameras.

\section{Related Work}
\label{sec:related_work}


In this section, we review the previous UAV-based human behavior datasets that are relevant to our dataset. Besides, since there is a lack of
\textit{large UAV-based} datasets for multi-modal behavior analysis, we also review some \textit{ground camera-based} multi-modal datasets.

\textbf{UAV-Based Human Behavior Understanding Datasets.}
Thanks to the flexibility, UAVs have been used in many scenarios where ground cameras may be difficult to be deployed, and some UAV-based benchmarks \cite{perera2018uav,barekatain2017okutama,perera2018uav,zhang2020person,ucfaa,ucfarg,oh2011large,mueller2016benchmark} have been introduced for human behavior understanding. However, to the best of our knowledge, all the existing benchmarks have limitations with regard to the dataset size, the diversities of scenes, the provided task categories, and captured data modality types, etc. 

Okutama-Action \cite{barekatain2017okutama} is a relatively small dataset for human action recognition, in which the RGB videos were collected
over a baseball field with UAVs. 
This dataset only includes 12 action classes performed by 9 subjects.
The small number of video samples and subjects, and the relatively simple scene obviously hinder its application for more challenging real-world scenarios. 


UAV-Gesture \cite{perera2018uav} is a dataset collected for UAV control gesture and pose analysis. 
This dataset provides 119 RGB video samples for 13 UAV control gestures, 
that were performed by 10 subjects with
relatively monotonous backgrounds. 

PRAI-1581 \cite{zhang2020person} is a UAV-based dataset for person Re-ID.
However this dataset only provides a single RGB modality and annotations for a single ReID task.


AVI \cite{singh2018eye} is a human behavior understanding dataset for violent action recognition and pose estimation.
Although this dataset provides annotations for two tasks,
this dataset is quite small and lacks diversities in multiple aspects. It contains 2K RGB images only.
Besides, only 5 action classes were performed by 25 subjects with small age differences (18 to 25).


Compared to all the existing UAV human behavior analysis datasets, our UAV-Human has significant advantages and provides many more videos and images, many more actions and poses, many more various scenes and backgrounds, much more diversified viewpoints and flying attitudes, many more data modalities, much more rich annotations, and many more tasks, etc. Thus our dataset shall be able to serve as a comprehensive and challenging benchmark for human behavior analysis with UAVs.



\textbf{Ground Camera-Based Multi-Modal Human Behavior Datasets.}
Since our Human-UAV dataset provides 6 different types of data modalities, here we also briefly review some of the ground camera-based human behavior datasets \cite{liu2017pku,ji2019large,liu2019ntu,rahmani2016histogram,hu2015jointly,kniaz2018thermalgan,kong2019mmact,martin2019drive,wang2019ev,kong2018human} that contain multi-modal data. 
SYSU 3DHOI \cite{hu2015jointly} is an indoor RGB-D dataset for human action recognition, which includes 480 video samples in 12 action categories, captured from 40 subjects. 
UWA3D Multiview II \cite{rahmani2016histogram} was captured from 10 subjects with 4 fixed camera angles for cross-view action understanding, which includes 1,075 videos and 30 action classes. 
PKU-MMD \cite{liu2017pku} contains 1,076 video sequences recorded in an indoor environment for human action recognition.
Varying-View RGB-D \cite{ji2019large} consists of 25,600 videos for action recognition, which were captured with ground-view cameras mounted on robots in the same indoor environment.
NTU RGB+D 120 \cite{liu2019ntu} is a large-scale dataset captured using three fixed cameras, which provides four data modalities including RGB, depth, IR and skeleton data. 

Unlike the aforementioned datasets that were captured using ground cameras, our UAV-Human is collected by flying UAVs with different speeds, heights, attitudes, and trajectories. Due to the flexibility of UAVs, our dataset provides unique viewpoints, obvious resolution variations, significant camera movements, and frequent motion blurs. Moreover, almost all the existing ground camera-based multi-modal human behavior datasets were collected under relatively simple, static, and monotonous scenes, while our dataset covers diverse outdoor and indoor scenes, different weather conditions, and various illumination and occlusion conditions, etc.

\textbf{Fisheye-Based Human Behavior Analysis Methods.}
Delibasis \textit{et al.} \cite{delibasis2014pose} proposed a deformable 3D human model to recognize the postures of a monitored person recorded by a fisheye camera. 
Srisamosorn \textit{et al.} \cite{srisamosorn2019human} introduced a histogram of oriented gradient descriptors to detect human body and head directions in fisheye videos. 
To handle distorted fisheye video samples,
here we propose to use the flat RGB images to guide a spatial transformer layer, which boosts the performance of action recognition in fisheye videos.

\begin{table*}
\caption{Comparisons among our UAV-Human dataset and some previous UAV-based human behavior analysis datasets, and ground camera-based multi-modal datasets. Our UAV-Human dataset significantly outperforms all the previous UAV-based human behavior datasets w.r.t. the data volume, the modalities, the diversities, and the labelled tasks and samples. Besides, it even obviously outperforms existing non-UAV (i.e., ground camera) based multi-modal datasets in lots of aspects.
}
\vspace{-0.4cm}
\scriptsize
\setlength\tabcolsep{3.8pt}
\centering
\begin{tabular}{!{\vrule width 1.0pt}c|c!{\vrule width 1.0pt}|c|c|c|c!{\vrule width 1.0pt}|c|c|c|c|c!{\vrule width 1.0pt}}

\noalign{\hrule height 1.0pt}

\multicolumn{2}{!{\vrule width 1.0pt}c!{\vrule width 1.0pt}|}{} & 
\multicolumn{4}{c!{\vrule width 1.0pt}|}{\textbf{Ground Camera-Based Multi-Modal Datasets}} & \multicolumn{5}{c!{\vrule width 1.0pt}}{\textbf{UAV-Based Datasets}} \\ 
\noalign{\hrule height 1.0pt}
\multicolumn{2}{!{\vrule width 1.0pt}c!{\vrule width 1.0pt}|}{Dataset Attribute} & 
\begin{tabular}[c]{@{}c@{}}PKU \\ MMD\end{tabular} &
\begin{tabular}[c]{@{}c@{}}Varying\\ View\end{tabular} &
\begin{tabular}[c]{@{}c@{}}NTU\\ RGBD120\end{tabular}
& \begin{tabular}[c]{@{}c@{}}Thermal \\ World\end{tabular} &
\begin{tabular}[c]{@{}c@{}} Okutama  \\ Action\end{tabular} &
\begin{tabular}[c]{@{}c@{}} UAV \\ Gesture\end{tabular} &
\begin{tabular}[c]{@{}c@{}} PRAI  \\ 1,581\end{tabular} &  AVI  &

\multicolumn{1}{c!{\vrule width 1.0pt}}{\begin{tabular}[c!{\vrule width 1.0pt}]{@{}c@{}} \textbf{UAV-Human} \\ \textbf{(Ours)}\end{tabular}}  \\                    
\noalign{\hrule height 1.0pt}

\multirow{3}{*}{Action Recog. Task} & \# Annotated Videos & 1,076 & 25,600 & 114,480 & $\times$ & 43 & 119 & $\times$ & NA & \textbf{22,476$\times$3} \\ \cline{2-11} 

& \# Annotated Subjects & 66 & 118 & 106 & $\times$ & 9 & 10 & $\times$ & 25 & \textbf{119} \\ \cline{2-11} 

& \# Annotated Classes & 51 & 40 & 120 & $\times$ & 12 & 13 & $\times$ & 5 & \textbf{155} \\ \hline

Pose Estimation Task & \# Annotated Samples & $\times$ & $\times$ & $\times$ & $\times$ & $\times$ & $\times$ & $\times$ & 2,000 & \textbf{22,476} \\ \hline

\multirow{2}{*}{Person ReID Task} & \# Annotated IDs & $\times$ & $\times$ & $\times$ & 516 & $\times$ & $\times$ & \textbf{1,581} & $\times$ & 1,144 \\ \cline{2-11}

& \# Annotated Samples & $\times$ & $\times$ & $\times$ & 15,118 & $\times$ & $\times$ & 39,461 & $\times$ & \textbf{41,290} \\ \hline

Attribute Recog. Task & \# Annotated Samples & $\times$ & $\times$ & $\times$ & $\times$ & $\times$ & $\times$ & $\times$ & $\times$ & \textbf{22,263} \\ 

\noalign{\hrule height 1.0pt}

\multirow{5}{*}{\begin{tabular}[c]{@{}c@{}}Data\\ Modality\end{tabular}} & RGB & $\checkmark$ & $\checkmark$ & $\checkmark$  & $\checkmark$ & $\checkmark$ & $\checkmark$ & $\checkmark$ & $\checkmark$ & \textbf{$\checkmark$} \\ \cline{2-11} 

& Depth & $\checkmark$ & $\checkmark$ & $\checkmark$ & $\times$ & $\times$ & $\times$ & $\times$ & $\times$ & \textbf{$\checkmark$} \\ \cline{2-11}

& IR & $\checkmark$ & $\times$ & $\checkmark$ & $\times$ & $\times$ & $\times$ & $\times$ & $\times$ & \textbf{$\checkmark$} \\ \cline{2-11} 

& Joint & $\checkmark$ & $\checkmark$ & $\checkmark$  & $\times$ & $\times$ & $\checkmark$ & $\times$ & $\checkmark$  & \textbf{$\checkmark$} \\ \cline{2-11} 

& Others & $\times$ & $\times$ & $\times$ & Thermal & $\times$ & $\times$ & $\times$ & $\times$ & \begin{tabular}[c]{@{}c@{}}\textbf{Fisheye} \\ \textbf{Night-vision}\end{tabular} \\ \hline
 
\multicolumn{2}{!{\vrule width 1.0pt}c!{\vrule width 1.0pt}|}{Sensors}  
& \begin{tabular}[c]{@{}c@{}} Kinect V2 \end{tabular}
& \begin{tabular}[c]{@{}c@{}} Kinect V2 \end{tabular} 
& \begin{tabular}[c]{@{}c@{}} Kinect V2 \end{tabular} 
& \begin{tabular}[c]{@{}c@{}} FLIR ONE  
\end{tabular}
& NA & GoPro 4 & NA & NA 
& \begin{tabular}[c]{@{}c@{}} \textbf{Azure DK,}\\ \textbf{Fisheye Camera,} \\ \textbf{Night-vision}  \textbf{Camera}\end{tabular} \\

\noalign{\hrule height 1.0pt}

\multirow{3}{*}{\begin{tabular}[c]{@{}c@{}}Capturing\\ Scenarios\end{tabular}} & \# Sites & 1 & 1 & 3 & 3 & 1 & 1 & 2 & 1 & \textbf{45} \\ \cline{2-11}

& Indoor & $\checkmark$ & $\checkmark$ & $\checkmark$ & $\checkmark$ & $\times$ & $\times$ & $\times$ & $\times$ & \textbf{$\checkmark$} \\ \cline{2-11} 

& Outdoor & $\times$ & $\times$ & $\checkmark$ & $\checkmark$ & $\checkmark$ & $\checkmark$ & $\checkmark$ & $\checkmark$ & \textbf{$\checkmark$} \\ \hline

\multirow{2}{*}{\begin{tabular}[c]{@{}c@{}}Challenging\\ Weathers\end{tabular}} & Windy & $\times$ & $\times$ & $\times$ & $\times$ & $\times$ & $\checkmark$ & $\times$ & $\times$ & \textbf{$\checkmark$} \\ \cline{2-11} 

& Rainy & $\times$ & $\times$ & $\times$ & $\checkmark$ & $\times$ & $\times$ & $\times$ & $\times$ & \textbf{$\checkmark$} \\ \hline

\multicolumn{2}{!{\vrule width 1.0pt}c!{\vrule width 1.0pt}|}{Complex Backgrounds} & $\times$ & $\times$ & $\checkmark$ & $\checkmark$ & $\times$ & $\times$ & $\times$ & $\times$ & \textbf{$\checkmark$} \\ \hline

\multicolumn{2}{!{\vrule width 1.0pt}c!{\vrule width 1.0pt}|}{Occlusion} & $\times$ & $\times$ & $\times$ & $\checkmark$ & $\checkmark$ & $\times$ & $\checkmark$ & $\times$ & \textbf{$\checkmark$} \\ \hline

\multicolumn{2}{!{\vrule width 1.0pt}c!{\vrule width 1.0pt}|}{Night Scenes} & $\times$ & $\times$ & $\times$ & $\checkmark$ & $\times$ & $\times$ & $\times$ & $\times$ & \textbf{$\checkmark$} \\ 

\noalign{\hrule height 1.0pt}

\multicolumn{2}{!{\vrule width 1.0pt}c!{\vrule width 1.0pt}|}{Camera Views} & fixed & varying & fixed & fixed & varying & varying & varying & varying & varying \\ \hline

\multirow{5}{*}{\begin{tabular}[c]{@{}c@{}}UAV \\ Attitudes\end{tabular}} & Hover & $\times$ & $\times$ & $\times$ & $\times$ & $\checkmark$ & $\checkmark$ & $\checkmark$ & \checkmark & \textbf{$\checkmark$} \\ \cline{2-11} 

& Lift & $\times$ & $\times$ & $\times$ & $\times$ & $\checkmark$ & $\times$ & $\times$ & $\times$ & \textbf{$\checkmark$} \\ \cline{2-11}

& Descent & $\times$ & $\times$ & $\times$ & $\times$ & $\checkmark$ & $\times$ & $\times$ & $\times$ & \textbf{$\checkmark$} \\ \cline{2-11}

& Cruising & $\times$ & $\times$ & $\times$ & $\times$ & $\times$ & $\times$ & $\checkmark$ & $\times$ & \textbf{$\checkmark$} \\ \cline{2-11}

& Rotating & $\times$ & $\times$ & $\times$ & $\times$ & $\times$ & $\times$ & $\checkmark$ & $\times$ & \textbf{$\checkmark$} \\ \cline{2-11}

\noalign{\hrule height 1.0pt}

\end{tabular}
\vspace{-0.2cm}
\end{table*}

\begin{figure}
    \centering
    \includegraphics[width=0.8\linewidth, height=1.2in, trim=0 2cm 0 1cm, clip]{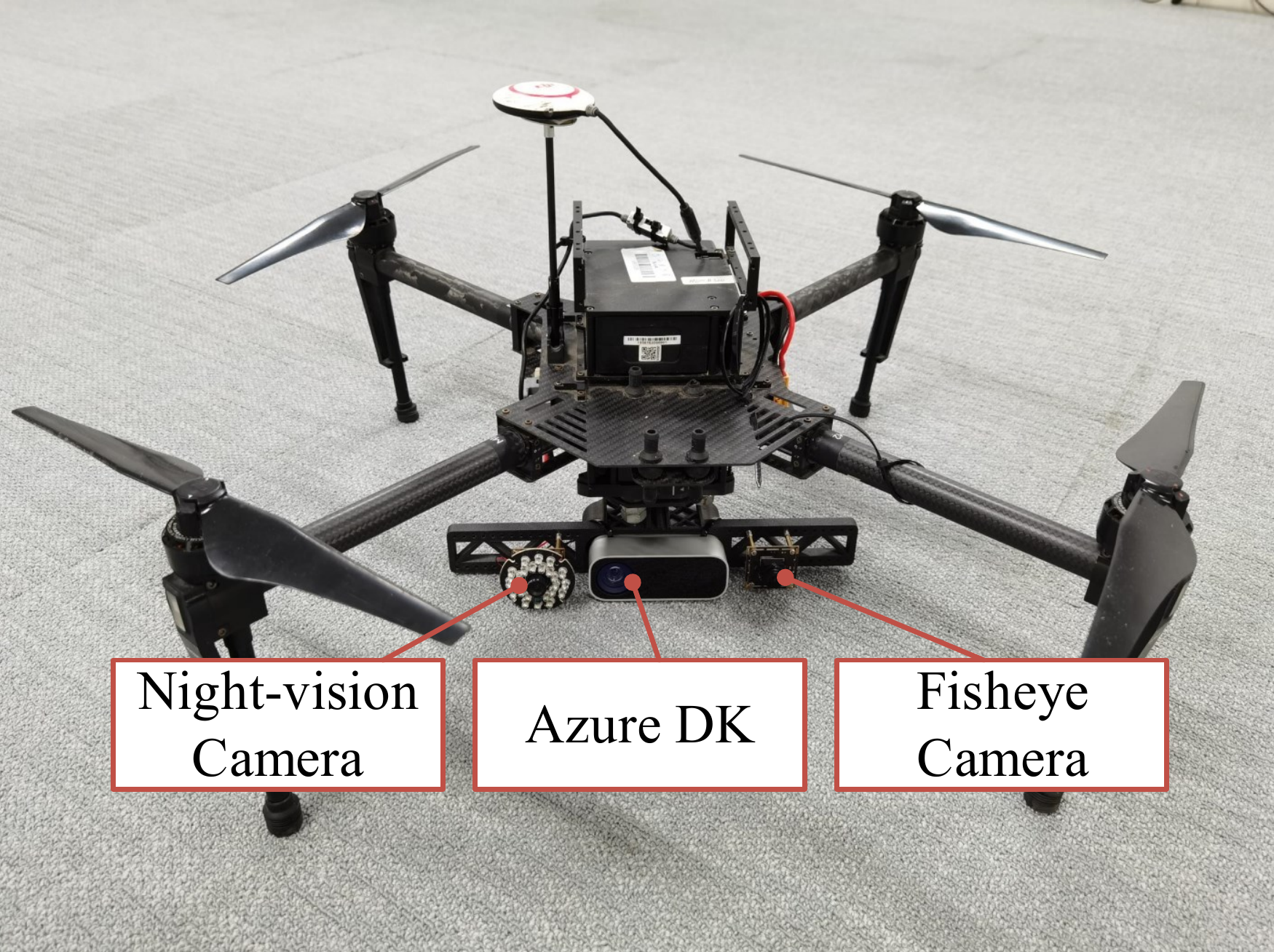}
    \vspace{-0.3cm}
    \caption{Illustration of our dataset collection platform deployed with multiple sensors.}
    \label{fig:platform}
    \vspace{-0.3cm}
\end{figure}

\section{UAV-Human Dataset}

\subsection{Dataset Description}
We create UAV-Human, a large-scale benchmark for UAV-based human behavior understanding,
which
contains 67,428 annotated video sequences of 119 subjects for action recognition, 
22,476 annotated frames for pose estimation, 
41,290 annotated frames of 1,144 identities for person re-identification, 
and 22,263 annotated frames for attribute recognition.
Our dataset is collected with a DJI Matrice 100 platform as shown in Figure~\ref{fig:platform}. 
To the best of our knowledge, UAV-Human is currently the largest, most challenging, and most comprehensive UAV dataset for human action, pose, and behavior understanding.
Some of the collected video samples are illustrated in Figure~\ref{fig:UAV-Human_samples}. 
Below we introduce the noteworthy features of our proposed dataset.

\textbf{Multiple Data Modalities.}
We collect multiple data modalities, of fisheye videos, night-vision videos, common RGB videos, infrared (IR) sequences, and depth maps, using a UAV with multiple sensors, including a fisheye camera, a night-vision sensor, and an Azure Kinect DK as shown in Figure~\ref{fig:platform}. Note that instead of using Microsoft Kinect V2 that has been widely used for indoor depth data collection \cite{liu2019ntu,shahroudy2016ntu}, we use the very recently released Azure Kinect DK to collect the IR and depth maps, since it is still consumer-available, yet is more powerful and more advanced, and can work better in both indoor and outdoor environments, compared to Microsoft Kinect V2. 
We actually observe the Azure Kinect DK works reliably and provides quite promising depth data in both indoor and outdoor scenarios during our dataset collection.

Besides the aforementioned modalities, we also
apply a pose estimator \cite{fang2017rmpe} on our dataset.
Hence the skeleton data is also provided for each video frame. 


Specifically, the RGB videos are recorded and stored at the resolution of 1920$\times$1080. The millimeter-measured depth maps are recorded in lossless compression formats at the resolution of $640\times576$.
The IR sequences are also stored at the resolution of $640\times576$.
The fisheye videos are captured by a 180$^\circ$-vision camera, and the resolution is $640\times480$.
The night-vision videos are recorded at two automatic modes, namely, color mode in the daytime and grey-scale mode in the nighttime, with the resolution at $640\times480$. 
The skeleton modality stores the positions of 17 major key-points of the human body.

The provided miscellaneous data modalities captured with different sensors that have different properties and different suitable application scenarios, shall be able to facilitate the community to exploit various methods to utilize different modalities, as well as cross-modality and modality-ensemble ones for UAV applications.

\textbf{Large Variations of Capture Environments.}
Our UAV-Human is captured from a total of 45 different sites across multiple rural districts and cities, which thus covers various outdoor and indoor scenes, including mountain areas, forests, river and lake-side areas, farmlands, squares, streets, gyms, university campuses, and several scenes inside buildings. Such abundant variations of scenes and backgrounds bring practical challenges to the UAV-based human behavior understanding problem.

\textbf{Long Time Period of Capturing.}
The overall time period of our dataset collection lasts for three months, across two different seasons (summer and fall), in both daytime and nighttime. Thus many distinctive features can be observed, such as the change of subject clothing fashions and surrounding styles, resulting from time period changes. In all, a long recording time period remarkably increases the diversities of the recorded video sequences.

\textbf{Various Challenging Weather Conditions.}
During our dataset collection, we encounter many adverse weather conditions including rainy and windy.
Specifically, occlusions caused by umbrellas, UAV shaking caused by strong wind, and low capture qualities caused by rain and fog, are all covered in our dataset.
Consequently, extreme weather conditions obviously lead to many challenging yet practical factors for our proposed dataset.

\textbf{Varying UAV Attitudes, Positions, and Views.}
In practical application scenarios, UAVs may have different flight attitudes, including hover, climb, descent, hovering turn, and side-ward flight, etc., which can result in significant camera shakes, 
affecting the locations of subjects in the captured frames, and leading to obvious motion blurs. Hence the diversified UAV flight attitudes and speeds in our dataset could encourage the community to develop robust models to handle human behavior 
analysis in such challenging scenarios. 
Besides, the hovering height (varying from $2$ to $8$ meters) in our dataset also brings large resolution variations of the subjects in the collected videos. 
Moreover, unlike many previous datasets \cite{perera2018uav}, where the capturing orientations are relatively constrained, our UAV-Human dataset provides very flexible UAV views.


\textbf{Multiple Human Behavior Understanding Tasks.}
For comprehensively analyzing human behaviors and actions from UAV videos, our proposed UAV-Human dataset provides annotations for four major tasks, namely, action recognition, pose estimation, person re-identification, and human attribute recognition, with very rich and elaborate annotations. 
Note that for all these tasks,
besides containing the universal features introduced above, each of them has the own notable contributions that are detailed in next section.

\subsection{Dataset Tasks and Annotations}
Based on the aforementioned data capturing platform and settings, we collect and annotate rich samples for the following different tasks.

\begin{figure}[ht]
    \centering
    \includegraphics[width=\linewidth]{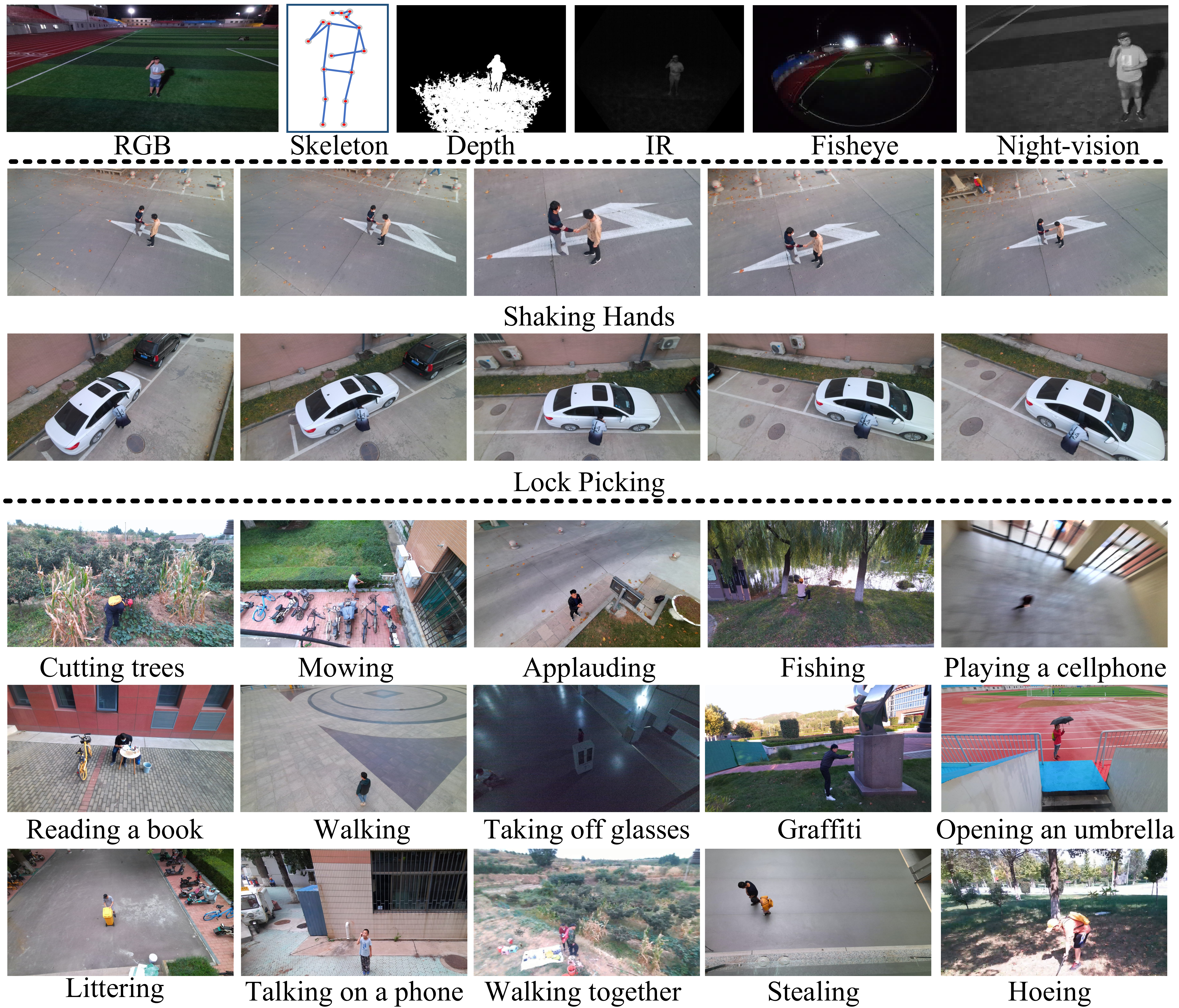}
    \vspace{-0.8cm}
    \caption{Examples of action videos in our UAV-Human dataset. The first row shows different data modalities. The second and third rows show two video sequences of significant camera motions and view variations, caused by continuously varying flight attitudes, speeds and heights. The last three rows display more action samples of our dataset, showing the diversities, e.g., distinct views, various capture sites, weathers, scales, and motion blur.
    }
    \label{fig:UAV-Human_samples}
    \vspace{-0.3cm}
\end{figure}

(1) \textbf{Action recognition with a large number of activity classes.}
To collect sufficient data for this task, we invite a total of 119 distinct subjects with different ages, genders, and occupations to naturally perform various actions and poses in all captured scenes during our dataset collection, which thus enables us to collect very diversified human actions,
yet still keep the action classes balanced.
Specifically, we collect 155 activity classes
in 6 different modalities (shown in Figure~\ref{fig:UAV-Human_samples}) covering different types of human behaviors that can be of great interest and significance for the practical UAV application scenarios. These activities include:
    \begin{inparaenum}[(i)]
        \item daily activities, e.g., smoking, wearing/taking off masks,
        \item productive activities, e.g., digging, fishing, mowing, cutting trees, carrying with shoulder poles, 
        \item violent activities, e.g., taking a hostage, stabbing with a knife, lock picking, 
        \item social interaction behaviors, e.g., walking together closely, whispering, 
        \item life-saving activities, e.g., calling for help, escaping, 
        and \item UAV control gestures.
    \end{inparaenum}
The aforementioned action classes can naturally occur in the wild, where CCTV surveillance cameras may be unavailable, while the UAVs can be used to flexibly track the performers of such activities in these scenarios. 

\begin{figure}[ht]
    \centering
    \includegraphics[width=0.85\linewidth, height=1.8cm]{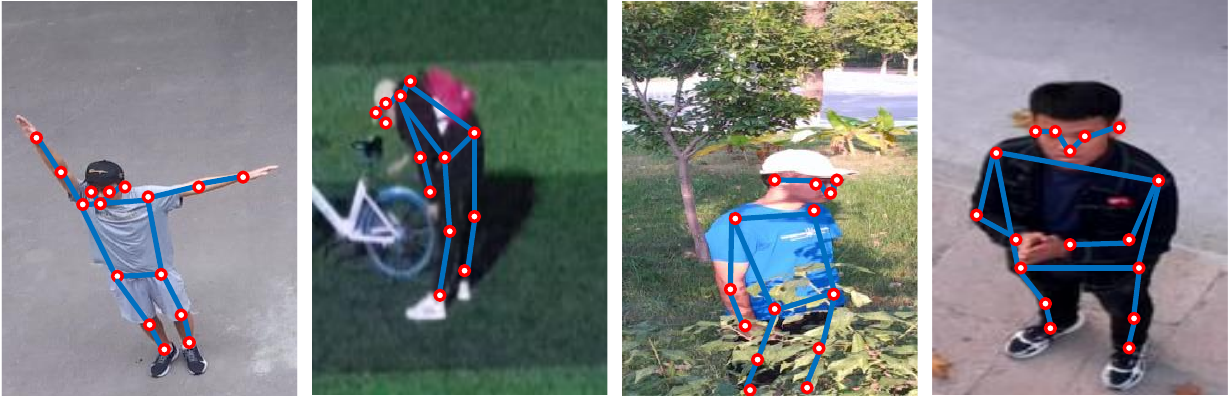}
    \vspace{-0.3cm}
    \caption{Samples for pose estimation.}
    \label{fig:pose_estimation}
    \vspace{-0.3cm}
\end{figure}

(2) \textbf{Pose estimation with manually-labeled human key-points.}
We invite 15 volunteers to label human poses in a total of 22,476 images that are sampled from the collected videos. 
For each image sample, a total of 17 major body joints are manually labelled, as illustrated in Figure~\ref{fig:pose_estimation}.
Note that the human pose estimation task in our dataset is quite challenging, 
owing to the distinct UAV viewpoints, different subject resolutions, diverse backgrounds, and various illumination, weather, and occlusion conditions.

\begin{figure}[ht]
    \centering
    \includegraphics[height=2cm,keepaspectratio]{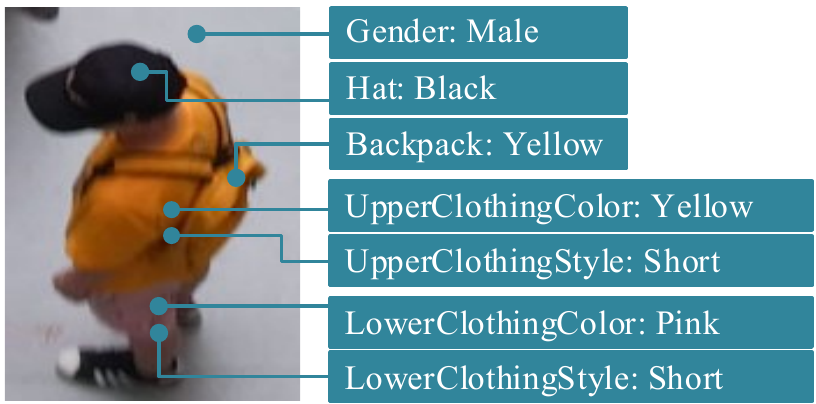}
    \vspace{-0.4cm}
    \caption{Samples for attribute recognition.}
    \label{fig:attribute}
    \vspace{-0.2cm}
\end{figure}

(3) \textbf{Attribute recognition with rich individual characteristics.}
The proposed dataset also provides 22,263 person images for human attribute recognition. We label 7 groups of attributes, including gender, hat, backpack, upper clothing color and style, as well as lower clothing color and style as shown in Figure~\ref{fig:attribute}. It is worth noting that the unconstrained views recorded from the UAV platform can cause large occlusions, leading to very challenging attribute recognition scenarios.

\begin{figure}[ht]
    \centering
    \includegraphics[width=\linewidth,height=1.5cm]{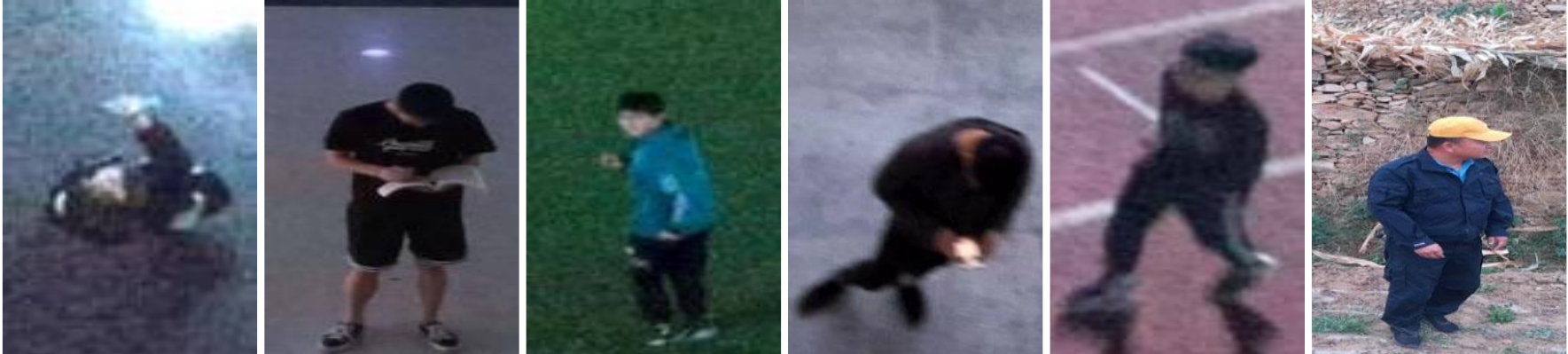}
    \vspace{-0.8cm}
    \caption{Samples for person re-identification.}
    \label{fig:re_id}
    \vspace{-0.3cm}
\end{figure}

(4) \textbf{ReID with various human poses and views.}
Besides, we collect 41,290 image samples with 1,144 identities, using our UAV platform
for person ReID. 
Captured from unique UAV perspectives, the person samples in this task also present very rich viewpoint, pose, and illumination variations, as shown in Figure~\ref{fig:re_id}.

We have obtained consent from all the captured human subjects to release all the tasks and samples in our dataset for non-commercial research and academic use.





\subsection{Evaluation Criteria}

\textbf{Action Recognition.}
We define two cross-subject evaluation protocols which are cross-subject-v1 (CSv1) and cross-subject-v2 (CSv2) for action recognition. For each evaluation protocol, we use 89 subjects for training and 30 subjects for testing. Note that for different cross-subject protocols, the IDs for training and testing subjects are different.
The classification accuracy is used for performance evaluation.

\textbf{Pose Estimation.}
We pick 16,288 frames from our manually-annotated frames for training and 6,188 frames for testing. Here mAP is used as the evaluation metric.

\textbf{Person Re-identification.}
For person re-identification, we use 11,805 images with 619 identities for training, 28,435 images with 525 identities for gallery, and the rest 1,050 images are query images. To measure the re-identification performance, we use the mean average precision (mAP) and cumulative match characteristic top-k accuracy (CMC\textsubscript{k}) as metrics. 


\textbf{Attribute Recognition.}
For attribute recognition, we use training and testing sets of 16,183 and 6,080 frames respectively.
To evaluate the performance of attribute recognition, we measure the classification accuracy for each attribute.

\begin{figure*}[ht]
    \centering
    \includegraphics[width=\linewidth]{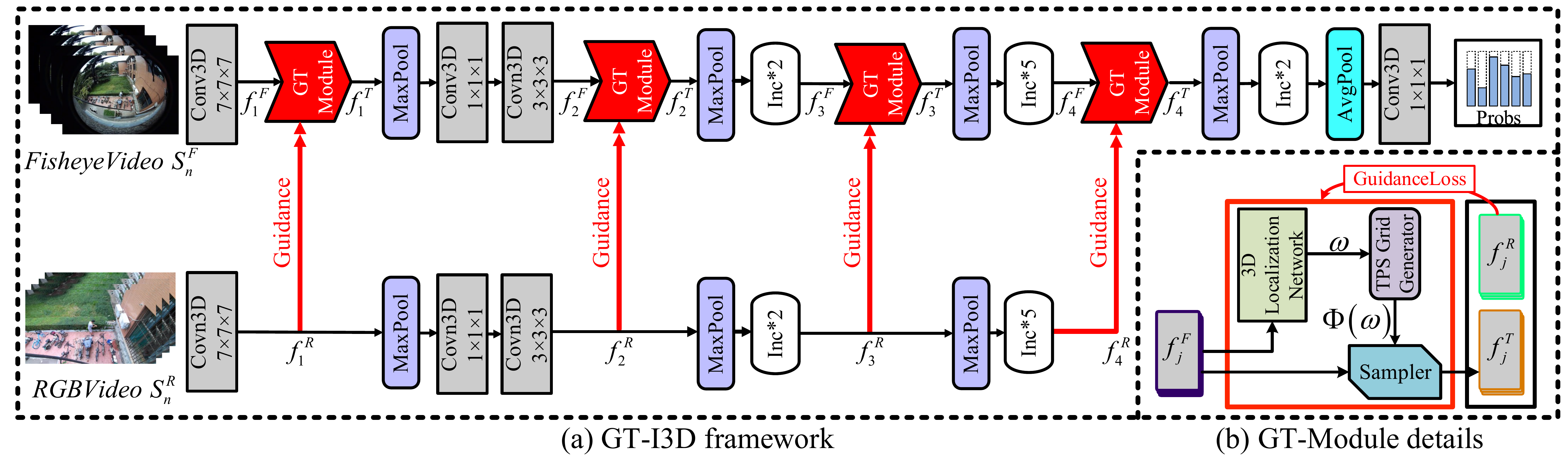}
    \vspace{-0.8cm}
    \caption{
    (a) Illustration of our Guided Transformer I3D (GT-I3D) framework. The proposed GT-I3D consists of a fisheye stream that learns to rectify the distorted fisheye video $S^{F}_{n}$ for action recognition, 
    by using integrated Guided Transformer Modules (GT-Modules) and a pre-trained RGB stream (fixed) fed with flat RGB video $S^{R}_{n}$ to guide the training of the GT-Modules. (b) Illustration of the detailed GT-Module. The 3D localization network is used to learn unbounded transformation parameters $\omega$ over the original fisheye features $f^{F}$. The grid generator is to derive the non-linear transformation $\Phi(\omega)$ from $\omega$. The sampler is used to warp the distorted fisheye features $f^{F}$ to the final transformed feature $f^{T}$ based on $\Phi(\omega)$. The obtained $f^{T}$ are further compared with the flat RGB features $f^{R}$ to constrain the GT-Module to learn better transformations.
    }
    \label{figs:overall_architecture}
    \vspace{-0.3cm}
\end{figure*}


\section{Guided Transformer Network for Fisheye Video Action Recognition}

Fisheye cameras are able to capture panoramic or hemispherical videos thanks to the ultra-wide view-angles
\cite{perera2018uav,barekatain2017okutama,singh2018eye}. Thus fisheye cameras are often equipped on UAVs to attain broad aerial views \cite{gurtner2009investigation}. However, videos captured by fisheye cameras are often distorted, 
which makes fisheye-based action recognition challenging.

In this section, we propose an effective method, named Guided Transformer I3D (GT-I3D)
to boost action recognition performance in fisheye videos as illustrated in Figure~\ref{figs:overall_architecture}.

\subsection{Guided Transformer I3D}

As shown in Figure~\ref{figs:overall_architecture}(a),
the overall architecture of our method consists of two streams, namely a fisheye stream and a pre-trained RGB stream.
We use I3D \cite{carreira2017quo} as our baseline network 
considering its powerful representation learning capabilities. However, fisheye video samples 
are always distorted
which brings difficulties for the original I3D to learn discriminative features.

Therefore, inspired by the spatial transformer network (STN) \cite{jaderberg2015spatial}, we propose a guided transformer module, GT-Module, as shown in Figure~\ref{figs:overall_architecture}(b). It can be inserted immediately before maxpooling layers in original I3D to warp each ``pixel'' on the feature maps extracted from distorted videos, by learning a series of unbounded transformations. The original STN \cite{jaderberg2015spatial} comprises a localization network learning transformation parameters
$\omega$ 
from source features, a grid generator deriving transformations 
$\Phi(\omega)$ 
from the learned transformation parameters, 
and a sampler learning to map source features to target features based on the transformations. 

Compared to original STN, we here use a 3D localization network replacing the original localization network, to learn 3D transformation parameters $\omega$ from fisheye video features $f^{F}$. Note that, to better alleviate the distortion problems, the learned parameters $\omega$ are unbounded, i.e. they are not normalized between -1 and 1 in our network. Then we use a grid generator to derive non-linear transformations $\Phi(\omega)$ conditioned on the learned parameters $\omega$. Finally, a sampler is employed to shift each pixel $(x_s, y_s)$ in the source fisheye feature maps $f^{F}$ to $(x_t, y_t)$ in the target feature maps $f^{T}$, as formulated in Eq.~\ref{eq:transformer}. Readers are also referred to the original paper of STN \cite{jaderberg2015spatial} for a detailed explanation of the mechanism.

\begin{equation}
    \vspace{-0.1cm}
    \left(\begin{array}{cc} x_t \\ y_t \end{array} \right) =
    \Phi(\omega)
    \left(\begin{array}{cc} x_s \\ y_s \end{array} \right)
    \label{eq:transformer}
\end{equation}

However, it is still difficult to train the original STN to learn how to effectively handle the distortion problems, because there lacks explicit guidance for the warping feature. Thus, we propose to use RGB videos in our UAV-Human dataset as the guidance information to constrain the learning of
the transformers by applying the Kullback–Leibler divergence loss between the flat RGB features $f^{R}$, and the transformed features $f^{T}$ as shown in Figure~\ref{figs:overall_architecture}.

It is worth noting that our GT-Module preserves the same shape among $f^{F}$, $f^{T}$ and $f^{R}$ 
with depth $D$, channel $C$, height $H$ and width $W$, i.e. $f^{F}$, $f^{T}$ and $f^{R}$ $\in \mathbb{R}^{D \times C \times H \times W}$. In addition, the learned transformations vary among the $D$ frames. This means if the input feature maps $f^{F}$ contain $D$ frames, the learned transformations also contain $D$ frames accordingly, and within each frame, we follow the original STN \cite{jaderberg2015spatial} to apply the same transformations to preserve spatial consistency across all channels of this frame.

\subsection{Training and Testing}
\textbf{Training.}
During training, the RGB and fisheye videos of the same action instance are respectively taken as inputs for the two streams. We uniformly sample $n$ frames from each video as the input ($S^{R}_{n}$ and $S^{F}_{n}$) for the RGB and fisheye streams respectively.
Note that the RGB stream has been pre-trained on the flat RGB video samples in advance, and thus the RGB stream is fixed here to guide the training of the fisheye stream. As mentioned above, all the GT-Modules are inserted immediately before maxpooling layers of I3D. Therefore, the fisheye features $f^{F}$ are passed through each GT-Module to achieve transformed features $f^{T}$ as shown in Figure~\ref{figs:overall_architecture}.
Meanwhile, the corresponding flat RGB features $f^{R}$ are used to guide each transformer by applying KL divergence constraint between $f^{T}$ and $f^{R}$ as formulated below,





\begin{equation}
    \vspace{-0.1cm}
    L_{G} = - \sum_{j=1}^J f_{j}^{R} \cdot (\log f_{j}^{T} - \log f_{j}^{R})
    \label{eq:kl_div}
\end{equation}

where $J$ is the number of inserted GT-Modules, i.e. the number of maxpooling layers in I3D,
as shown in Figure~\ref{figs:overall_architecture}. In addition, a standard classification loss is used to update the fisheye stream by comparing the predicted score $\hat{y}_k$ and the class label $y_k$ as follows,

\begin{equation}
    \vspace{-0.1cm}
    L_{C} = -\sum_{k=1}^{K} y_k \log \hat{y}_k
    \label{eq:entropy}
\end{equation}

where $K$ is the number of action classes. In summary, the following overall objective function
can be used to update our GT-I3D framework.

\begin{equation}
    \vspace{-0.1cm}
    L = L_{G} + L_{C}
    \label{eq:all_loss}
\end{equation}

\textbf{Testing.}
During evaluation, only the fisheye video sequence $S^{F}_{n}$ need to be sent to the fisheye stream to predict the action category, while the RGB stream is not required. Thus for testing scenarios, if the UAV is only deployed with a fisheye camera, we can still use this well-trained fisheye stream for more reliable action recognition without requiring flat RGB videos.


\textbf{Implementation Details.}
To train our GT-I3D framework, the initial learning rate is set to 4e-3
and batch size is set to 16. For $S^{R}_{n}$ and $S^{F}_{n}$, we choose $n=64$ frames as the inputs. We use SGD as the optimizer to train our network on 2 Nvidia RTX2080Ti for 30K iterations.



\section{Experiments}
\subsection{Evaluation on Action Recognition}

\textbf{Evaluation of multiple modalities.}
The proposed UAV-Human provides multiple data modalities for action recognition.
Therefore we evaluate the performance of I3D network \cite{carreira2017quo} on different data modalities and report action recognition results
in Table~\ref{tab:other_modalities}.

We can observe that when using night-vision and IR videos as inputs, the highest accuracies of $28.72\%$ and $26.56\%$ are achieved on CSv1. This is partially because a large portion of videos are collected in dark environments, while night-vision and IR videos attain clearer visual information in the nighttime compared to other modalities.
Note that performance of depth sequences is slightly weaker than RGB videos mainly because depth sequences contain noises.
Moreover, the performance discrepancy between fisheye videos and RGB videos is mainly caused by 
the intrinsic distortion issue of fisheye sensors.

Our proposed method is able to rectify distorted fisheye videos and thus improves the performance of original fisheye methods.
The performance of our full model ($23.24\%$) is even competitive to the RGB model ($23.86\%$) on CSv1 as shown in Table~\ref{tab:other_modalities}. 
Moreover, ablation studies are also conducted. 

First we use guidance loss only, by applying KL divergence constraint directly on two feature maps encoded by the fisheye and RGB streams respectively, and attain performance of $21.68\%$ on CSv1. Then we use video transformers
only, without guidance information from the RGB stream, and obtain performance of $21.49\%$ on CSv1. We can observe that
our full model also achieves the highest accuracy among all fisheye-based methods.
This indicates the efficacy of our proposed GT-Module. 

Note that, for RGB videos we also compare the performances of I3D baseline \cite{carreira2017quo} and TSN baseline \cite{wang2018temporal}, and the recognition accuracies (CSv1) are $23.86\%$ and $18.15\%$ respectively. 


\textbf{Evaluation of the state-of-the-art methods on skeleton.}
Here we evaluate different state-of-the-art methods \cite{yan2018spatial,shi2019two,shi2019skeleton,li2020hardnet,cheng2020skeleton} on skeleton modality of the UAV-Human. As shown in Table~\ref{tab:skeleton_modality}, 
all skeleton-based methods outperform video-based methods shown in Table~\ref{tab:other_modalities}.
The reason is that videos are sensitive to varying backgrounds, scales and locations led by the continuously changing of camera positions and viewpoints of the UAV.
However, skeletal representations are more robust to complex backgrounds and additional normalization methods such as scaling and translating the skeletons \cite{liu2019ntu}
can be applied,
which can make skeleton a more robust representation compared to other modalities in such a challenging UAV scenario. 

\begin{table}[]
    \footnotesize
    \caption{Evaluation of using different modalities for action recognition. All modalities achieve relatively low accuracies due to the practical challenges in our UAV-Human dataset, such as continuously changing of views, scales and locations and blurry images caused by different UAV flight attitudes, speeds and heights.}
    \vspace{-0.4cm}
    \centering
    \begin{tabularx}{\linewidth}{lcc}
        \toprule
         Modality & \makecell{CSv1 \\ Accuracy (\%)}  & \makecell{CSv2 \\ Accuracy (\%)} \\
         \noalign{\hrule height 1pt}
         RGB Video & 23.86 & 29.53\\
         \hline
         Depth Video & 22.11 & -\\
         \hline
         IR Video & 26.56 & -\\
         \hline
         Night-vision Video & 28.72 & -\\
         \hline
         Fisheye Video & 20.76 & 34.12\\
         \hline
         \hline
         Fisheye Video+Guidance Loss & 21.68 & -\\
         \hline
         Fisheye Video+Video Transformer & 21.49 & -\\
         \noalign{\hrule height 1pt}
         Fisheye Video+Our Full Model & \textbf{23.24} & -\\
         \bottomrule
    \end{tabularx}
    \label{tab:other_modalities}
    \vspace{-0.25cm}
\end{table}

\begin{table}[]
    \footnotesize
    \caption{The results of skeleton-based action recognition.}
    \vspace{-0.4cm}
    \centering
    \begin{tabularx}{0.75\linewidth}{lcc}
        \toprule
         Method & \makecell{CSv1 \\ Accuracy (\%)}  & \makecell{CSv2 \\ Accuracy (\%)} \\
         \noalign{\hrule height 1pt}
         DGNN\cite{shi2019skeleton} & 29.90 & -\\
         \hline
         ST-GCN \cite{yan2018spatial} & 30.25 & 56.14\\
         \hline
         2S-AGCN \cite{shi2019two} & 34.84 & 66.68\\
         \hline
         HARD-Net \cite{li2020hardnet} & 36.97 & -\\
         \hline
         Shift-GCN \cite{cheng2020skeleton} & 37.98 & 67.04\\
         \bottomrule
    \end{tabularx}
    \label{tab:skeleton_modality}
    \vspace{-0.5cm}
\end{table}

\subsection{Evaluation on Pose Estimation}
Table~\ref{tab:pose_estimation} shows the results of two prevalent pose estimation methods \cite{fang2017rmpe,cheng2020higherhrnet} on the UAV-Human dataset. 
We can observe that both methods attain relatively weak performance ($56.9\%$ \cite{cheng2020higherhrnet} and $56.5\%$ \cite{fang2017rmpe} respectively) on our UAV-Human dataset. This is possibly because that varying scales and views caused by multiple UAV attitudes plus diversified subjects' postures and complex occlusions bring more challenges to pose estimation in UAV application scenarios.


\begin{table}[]
\footnotesize
    \caption{The results of pose estimation.}
    \centering
    \vspace{-0.4cm}
    \begin{tabularx}{0.8\linewidth}{l>{\centering\arraybackslash}X}
        \toprule
         Method & mAP(\%) \\
         \noalign{\hrule height 1pt}
         HigherHRNet \cite{cheng2020higherhrnet} & 56.5\\
         \hline
         AlphaPose \cite{fang2017rmpe} & 56.9 \\
         \bottomrule
    \end{tabularx}
    \label{tab:pose_estimation}
    \vspace{-0.2cm}
\end{table}

\subsection{Evaluation on Person Re-Identification}

Three state-of-the-art person ReID methods \cite{suh2018part,sun2018beyond,luo2019bag,zheng2019joint} are evaluated on our dataset. 
The results are presented in Table~\ref{tab:ReID}. 
Note our dataset is collected by moving cameras on a UAV, and the person images are often captured from overhead perspectives.
This means our dataset brings brand new challenges to person re-identification, that can
encourage the future deep neural networks to learn more representative features.
 


\begin{table}[]
\footnotesize
    \caption{Results of person re-identification.}
    \vspace{-0.4cm}
    \centering
    \begin{tabular}{lccc}
        \toprule
         Method & mAP & Rank-1 & Rank-5  \\
         \noalign{\hrule height 1pt}
         Part-Aligned \cite{suh2018part} & 60.86 & 60.86 & 81.71 \\ 
         \hline
         PCB \cite{sun2018beyond} & 61.05 & 62.19 & 83.90 \\ 
         \hline
         Tricks \cite{luo2019bag} & 63.41 & 62.48 & 84.38 \\ 
         \hline
         DG-Net \cite{zheng2019joint} & 61.97 & 65.81 & 85.71 \\
         \bottomrule
    \end{tabular}
    \label{tab:ReID}
    \vspace{-0.2cm}
\end{table}

\begin{table}[]
\footnotesize
    \caption{Results of Attribute Recognition. UCC/S and LCC/S represent Upper Clothing Color/Style and Lower Clothing Color/Style respectively.}
    \vspace{-0.4cm}
    \centering
    \begin{tabularx}{\linewidth}{l>{\centering\arraybackslash}Xcccc}
        \toprule
         \multirow{2}{*}{Method} & \multicolumn{5}{c}{Accuracy (\%)}\\
         \cline{2-6}
         & Gender & Hat & Backpack & UCC/S & LCC/S \\
         \noalign{\hrule height 1pt}
         ResNet \cite{he2016deep} & 74.7 & 65.2 & 63.5 & 44.4/68.9 & 49.7/69.3 \\
         \hline
         DenseNet \cite{huang2017densely} & 75.0 & 67.2 & 63.9 & 49.8/73.0 & 54.6/68.9 \\
         \bottomrule
    \end{tabularx}
    \label{tab:attribute_recognition}
    \vspace{-0.5cm}
\end{table}



\subsection{Evaluation on Attribute Recognition}

We train two baseline methods using ResNet and DenseNet models as their respective feature extractors to identify common visual attributes. 
As shown in Table~\ref{tab:attribute_recognition},
recognition on clothing colors and styles achieve the lowest accuracies. This is possibly because that our dataset is captured in a relatively long period of time, and thus we have diversified subjects with different dressing types, plus large variations of viewpoints caused by multiple UAV attitudes, making our UAV-Human a challenging dataset for attribute recognition.

\section{Conclusion}
To the best of our knowledge, our dataset is the largest, most challenging and most comprehensive UAV-based dataset for human action, pose, and behavior understanding. We believe that the proposed UAV-Human will encourage the exploration and deployment of various data-intensive learning models for UAV-based human behavior understanding. We also propose a GT-I3D network for distorted fisheye video action recognition. The experimental results show the efficacy of our method.


\bibliographystyle{ieee_fullname}
\bibliography{ref}

\end{document}